\pdfoutput=1

\documentclass[11pt]{article}

\usepackage[]{emnlp2021}

\usepackage{times}
\usepackage{latexsym}

\usepackage[T1]{fontenc}
\usepackage[utf8]{inputenc}

\usepackage{microtype}
\usepackage{pifont}
\newcommand{\xmark}{\ding{55}}%
\usepackage{amsmath}
\usepackage{amssymb}
\usepackage{graphicx}
\usepackage{subfig}
\usepackage{placeins}
\usepackage{float}
\usepackage{subfig}
\usepackage{enumitem}
\usepackage{adjustbox}
\usepackage{commath}
\graphicspath{ {./images/} }

\usepackage{bbm}
\usepackage{multirow}
\usepackage{caption}
\usepackage{array}
\usepackage{xcolor}
\definecolor{myOrange}{rgb}{1,0.5,0}

\newcolumntype{R}[2]{%
    >{\adjustbox{angle=#1,lap=\width-(#2)}\bgroup}%
    l%
    <{\egroup}%
}

\makeatletter
\newcommand{\thickhline}{%
    \noalign {\ifnum 0=`}\fi \hrule height 1pt
    \futurelet \reserved@a \@xhline
}

\usepackage{xcolor}
\definecolor{outerspace}{RGB}{68, 80, 70}
\definecolor{OLIVINE}{RGB}{147, 184, 115}
\definecolor{RED}{RGB}{255, 0, 0}
\definecolor{BLUE}{RGB}{0,0,255}
\title{Visual Goal-Step Inference using \textcolor{outerspace}{wiki}\textcolor{OLIVINE}{How}}

\author{Yue Yang,  Artemis Panagopoulou, Qing Lyu, Li Zhang,\\ \textbf{Mark Yatskar, Chris Callison-Burch}\\
Department of Computer and Information Science, University of Pennsylvania\\
{\small{\tt \{yueyang1,artemisp,lyuqing,zharry,myatskar,ccb\}@seas.upenn.edu}}
}

\date{}

\begin{document}
\maketitle
\begin{abstract}
Understanding what sequence of steps are needed to complete a goal can help artificial intelligence systems reason about human activities. Past work in NLP has examined the task of goal-step inference for text. We introduce the visual analogue. We propose the Visual Goal-Step Inference (VGSI) task, where a model is given a textual goal and must choose which of four images represents a plausible step towards that goal. With a new dataset harvested from wikiHow consisting of 772,277 images representing human actions, we show that our task is challenging for state-of-the-art multimodal models. Moreover, the multimodal representation learned from our data can be effectively transferred to other datasets like HowTo100m, increasing the VGSI accuracy by 15 - 20\%. Our task will facilitate multimodal reasoning about procedural events.
\end{abstract}

\section{Introduction}
Recently, there has been growing attention on the representation of complex events, with renewed interest in script learning and commonsense reasoning \cite{park2018learning, mujtaba2019recent, li2020connecting}. One aspect of event representation is the relationship between high-level goals and the steps involved \cite{lyu-zhang-wikihow:2020, zhang-etal-2020-intent}. For example, given a goal (e.g. ``change a tire''), an intelligent system should be able to infer what steps need to be taken to accomplish the goal (e.g. ``place the jack under the car'', ``raise the jack''). 
In most work, events are represented as text \cite{zellers2018swagaf, coucke2018snips, zhang-etal-2019-joint}, while they could have different modalities in the real world.

Learning \textit{goal-step relations} in a multimodal fashion is an interesting challenge since it requires reasoning beyond image captioning. We contend that multimodal event representation learning will have interesting implications for tasks such as schema learning \cite{li2020connecting, li2021future} to mitigate reporting bias \cite{gordon2013reporting} since steps are often not explicitly mentioned in a body of text. For instance, if a robot is asked to ``get a slice of cake,''  it has to know that it should ``take the cake out of the box'', ``cut a slice'', ``put it on a plate'', and then ``take the plate to the user''. Such steps are commonsense to people and thus rarely specified explicitly, making them hard to infer from textual data. However, with multimodal learning, we could obtain such details from visual signals. This multimodal goal-step relation could also be used for vision-enabled dialog systems\footnote{Like the \href{https://www.amazon.science/academic-engagements/amazon-launches-new-alexa-prize-taskbot-challenge}{Alexa Prize Taskbot Challenge}.} to recognize what task a user is completing and provide helpful recommendations. 

\begin{figure}[!t]
\centering
    \includegraphics[width=\linewidth]{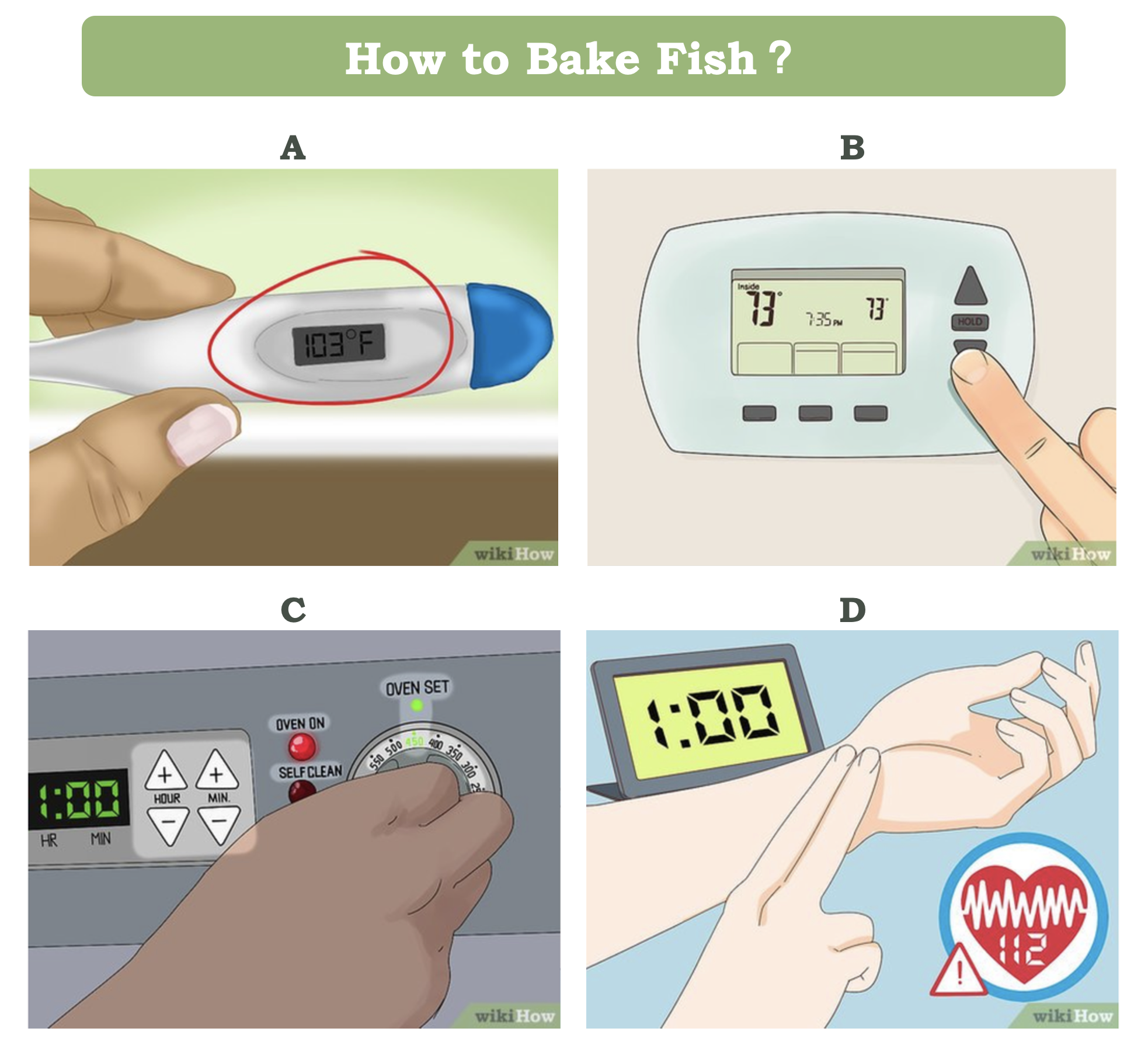}
    \captionsetup{justification=centering}
    \caption{An example Visual Goal-Step Inference Task: given a text goal ({\it bake fish}), select the image (C) that represents a step towards that goal.}
    \label{fig:example}
\end{figure}

We propose a new task called \textbf{Visual Goal-Step Inference (VGSI)}: given a textual goal and multiple images representing candidate events, a model must choose one image which constitutes a reasonable step towards the given goal. This means that a model should correctly recognize not only the specific action illustrated in an image (e.g., ``turning on the oven'', in Figure~\ref{fig:example}), but also the intent of the action (``baking fish'').

\begin{figure}[!t]
\centering
    \includegraphics[width=7.5cm]{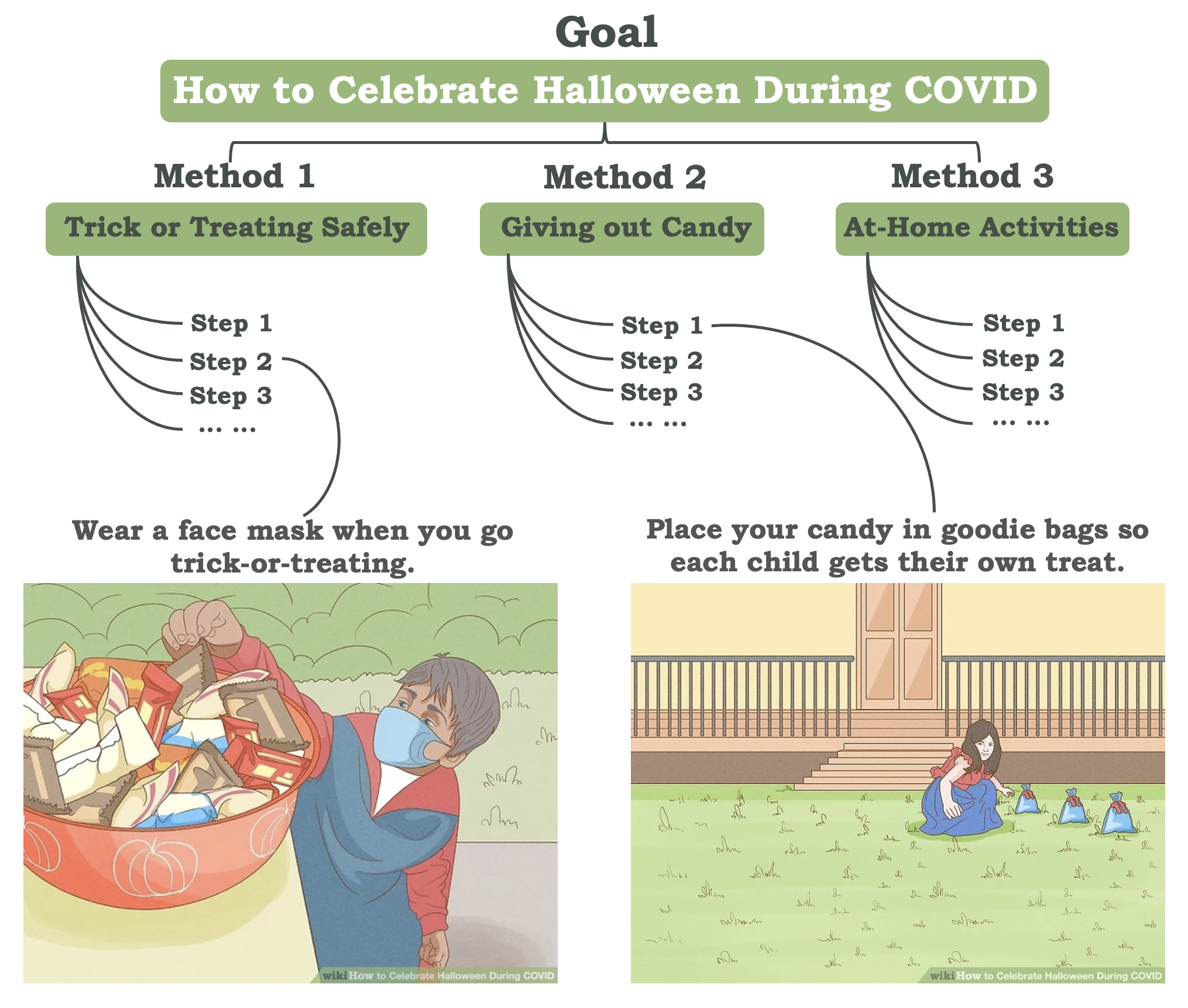}
    \caption[semantic relationship]{Hierarchical multimodality of wikiHow.}
    \label{fig:semantical}
    \vspace{-0.15in}
\end{figure}

We collect data from \href{www.wikihow.com}{\textcolor{outerspace}{\textbf{wiki}}\textcolor{OLIVINE}{\textbf{How}}} articles, where most steps are illustrated with images. Our VGSI training set is constructed using three sampling strategies to select negative image candidates as distractors. In the format of multiple-choice and image retrieval, we evaluate four state-of-the-art multimodal models: DeViSE \cite{frome2013devise}, Similarity Networks \cite{wang2018learning}, Triplet Networks  \cite{hoffer2015deep}, and LXMERT \cite{tan2019lxmert} to human performance. It is observed that SOTA models designed for caption-based multimodal tasks \cite{karpathy2014deep, johnson2016densecap} struggle on VGSI, exhibiting a 40\% gap in accuracy from human performance when using a challenging sampling strategy.

One limitation of wikiHow is that most images are drawings rather than photos (which are more typically used in computer vision research). The knowledge learned from wikiHow is nevertheless useful when applied to real photos. We demonstrate this by pre-training a triplet-network on our wikiHow VGSI task and then conducting transfer learning on out-of-domain datasets. Our experiments show that pre-trained models can effectively transfer the goal-step knowledge to task-oriented video datasets, such as COIN \cite{Tang_2019_CVPR} and Howto100m \cite{miech2019howto100m}. In addition, we design an aggregation model on top of SOTA models which treats wikiHow as a knowledge base that further increases the transfer learning performance (see Appendix~\ref{step_agg}).

We make three key contributions: (1) We propose the VGSI task, which is more challenging than traditional caption-based image-text matching tasks and requires the model to have an intermediate reasoning process about goal-step relations.
(2) To study the VGSI task, we collect a multimodal dataset from wikiHow which contains over 770k images.
(3) Through transfer learning, we show that the knowledge learned from our dataset can be readily applied to out-of-domain datasets, with an accuracy improvement of 15-20\% on VGSI.

\section{wikiHow as Multimodal Resource}
We use wikiHow as the data source for VGSI because it has been successfully adopted in prior work for procedural learning \cite{zhou-etal-2019-learning-household} and intent detection \cite{zhang-etal-2020-intent} in the language domain. As shown in Figure~\ref{fig:semantical}, each wikiHow article contains a high-level \emph{goal} and one or more different \emph{methods}\footnote{In some \href{https://www.wikihow.com/Make-a-Shadow-Box}{articles}, they use \emph{parts} instead of \emph{methods}.} to achieve it. Each method then includes a series of specific \emph{steps}, typically accompanied with corresponding images.

\begin{table}[!t]
\centering
\resizebox{7.5cm}{!}{%
\begin{tabular}{ccccc}
\thickhline
\textbf{Category}                                                              & \textbf{Goals}                      & \textbf{Methods }                    & \textbf{Steps}                       & \textbf{Images }                     \\ \hline
Health                                                                 & 7.8k                      & 19.1k                      & 97.5k                      & 111.8k                     \\
Home and Garden                                                        & 5.9k                      & 16.0k                      & 82.9k                     & 85.4k                      \\
\begin{tabular}[c]{@{}c@{}}Education \& \\ Communications\end{tabular} & 4.7k                      & 12.4k                      & 61.2k                      & 66.1k                      \\
Food \& Entertaining                                                   & 4.6k                      & 11.6k                      & 62.0k                      & 69.0k                      \\
Finance \& Business                                                    & 4.4k                      & 11.8k                      & 59.3k                      & 66.8k                      \\
Pets \& Animals                                                        & 3.5k                      & 9.5k                       & 45.3k                      & 48.0k                      \\
Personal Care \& Style                                                 & 3.4k                      & 9.0k                       & 46.1k                     & 48.9k                    \\
Hobbies \& Crafts                                                      & 2.8k                      & 7.5k                       & 40.9k                      & 42.7k                      \\
Computers \& Electronics                                               & 2.6k                      & 6.1k                       & 31.5k                      & 36.2k                      \\
Arts \& Entertainment                                                  & 2.5k                      & 6.8k                       & 35.4k                      & 37.2k                      \\ \hline
Total                                                                  & \multicolumn{1}{l}{53.2k} & \multicolumn{1}{l}{155.3k} & \multicolumn{1}{l}{772.3k} & \multicolumn{1}{l}{772.3k} \\ \thickhline
\end{tabular}
}
\captionsetup{justification=centering}
\caption{Number of goals, methods, steps and images in the top 10 wikiHow categories.}\vspace{-0.5cm}
\label{category}
\end{table}

The format of wikiHow articles provides a hierarchical multimodal relationship between images and sentences. We can obtain three types of text-image pairs from wikiHow, in increasing specificity: goal-image, method-image, and step-image. However, these text-image pairs are not enough information for a system to succeed on VGSI; it also needs the appropriate background knowledge. For the example in Figure~\ref{fig:semantical}, the system needs to know that ``Trick-or-Treating'' and ``candies'' are Halloween traditions and that a ``mask'' is required during ``COVID-19''.

In total, as shown in Table~\ref{category}, the corpus consists of 53,189 wikiHow articles across various categories of everyday tasks, 155,265 methods, and 772,294 steps with corresponding images \footnote{Both datasets and code are available \href{https://github.com/YueYANG1996/wikiHow-VGSI}{here}.}

\section{Methods}
\subsection{Problem Formulation}
Given a high-level goal $G$---defined as a sequence of words---and an image $I \in \mathbb{R}^{3 \times h \times w}$---with the dimension of 3 color channels, the width, and the height---the model outputs the matching score:
\begin{equation} \label{eq1}
    match(G, I) = F (X_G, X_I)
\end{equation}
in which, $X_G \in \mathbb{R}^{d_G} $ and $X_I \in \mathbb{R}^{d_I}$ are the feature representations of the goal and the image, respectively. $F$ is the scoring function that models the interactions between the two representations. At inference time, the model will choose the candidate with the highest matching score as the prediction.

\subsection{Models}

\noindent
\textbf{DeViSE} takes in the pre-trained embedding vectors from the two modalities and maps the source vector onto the span of the target vector. DeViSE is trained only on the positive pairs $(G, I)$  and projects an image embedding onto the same dimension as the goal with L2 normalization. Then it computes the cosine similarity of the two normalized vectors as the matching score.

\noindent
\textbf{Similarity Network} Each branch of the similarity network maps one modality to the joint span and executes point-wise multiplication to construct a joint vector. The last layer is fully-connected with softmax activation and outputs an array of size two to denote the weights of each class for binary classification. We compute the matching score by taking the dot product $[1, -1]$ with the output.

\noindent
\textbf{Triplet Network} requires the input to be the format of a triplet $(G, I_{pos}, I_{neg})$. Three branches in the network map the three embeddings to the same joint span, such that the branches of positive and negative images share the same weight. The network learns the cross-modality by minimizing the positive pair distance and maximizing the negative pair distance. We choose cosine distance as the metric which is also used as the matching score.

\noindent
\textbf{LXMERT} is a multimodal encoder that aims to ground text to images through attention layers. The image input is represented as a sequence of objects and the sentence input is a sequence of words. LXMERT utilizes two single-modality transformer encoders (language and object encoders) and a cross-modality transformer encoder to achieve the attention mechanism and capture the relationship between the two modalities. Same as the similarity network, LXMERT is trained as a binary classifier.

\begin{table}[!t]
\centering
\resizebox{7.5cm}{!}{%
\begin{tabular}{c|ccc}
\thickhline
\multirow{2}{*}{\textbf{Model}} & \multicolumn{3}{c}{\textbf{Sampling Strategy} (Test Size)}                                                                                                                                         \\ \cline{2-4} 
                       & \begin{tabular}[c]{@{}c@{}}\textbf{Random}\\ (153,961)\end{tabular} & \begin{tabular}[c]{@{}c@{}}\textbf{Similarity}\\ (153,770)\end{tabular} & \begin{tabular}[c]{@{}c@{}}\textbf{Category}\\ (153,961)\end{tabular} \\ \hline
Random                 & .2500                                                     & .2500                                                         & .2500                                                       \\
DeViSE                 & .6719                                                     & .3364                                                         & .4558                                                       \\
Similarity Net         & .6895                                                     & .6226                                                         & .4983                                                       \\
LXMERT                 & .7175                                                     & .4259                                                         & .2886                                                       \\
Triplet Net (GloVe)    & .7251                                                     & .7450                                                         & .5307                                                       \\
Triplet Net (BERT)     & \textbf{.7280}                                            & \textbf{.7494}                                                & \textbf{.5360}                                              \\ \hline
Human                  & .8450                                                     & .8214                                                         & .7550  \\ \thickhline
\end{tabular}
}
\caption{Accuracy of SOTA models on the wikiHow VGSI test set with different sampling strategies (sample size is shown in parentheses).}
\label{indomain model}
\end{table}

\begin{figure}[!t]
\centering
    \includegraphics[width=7.5cm]{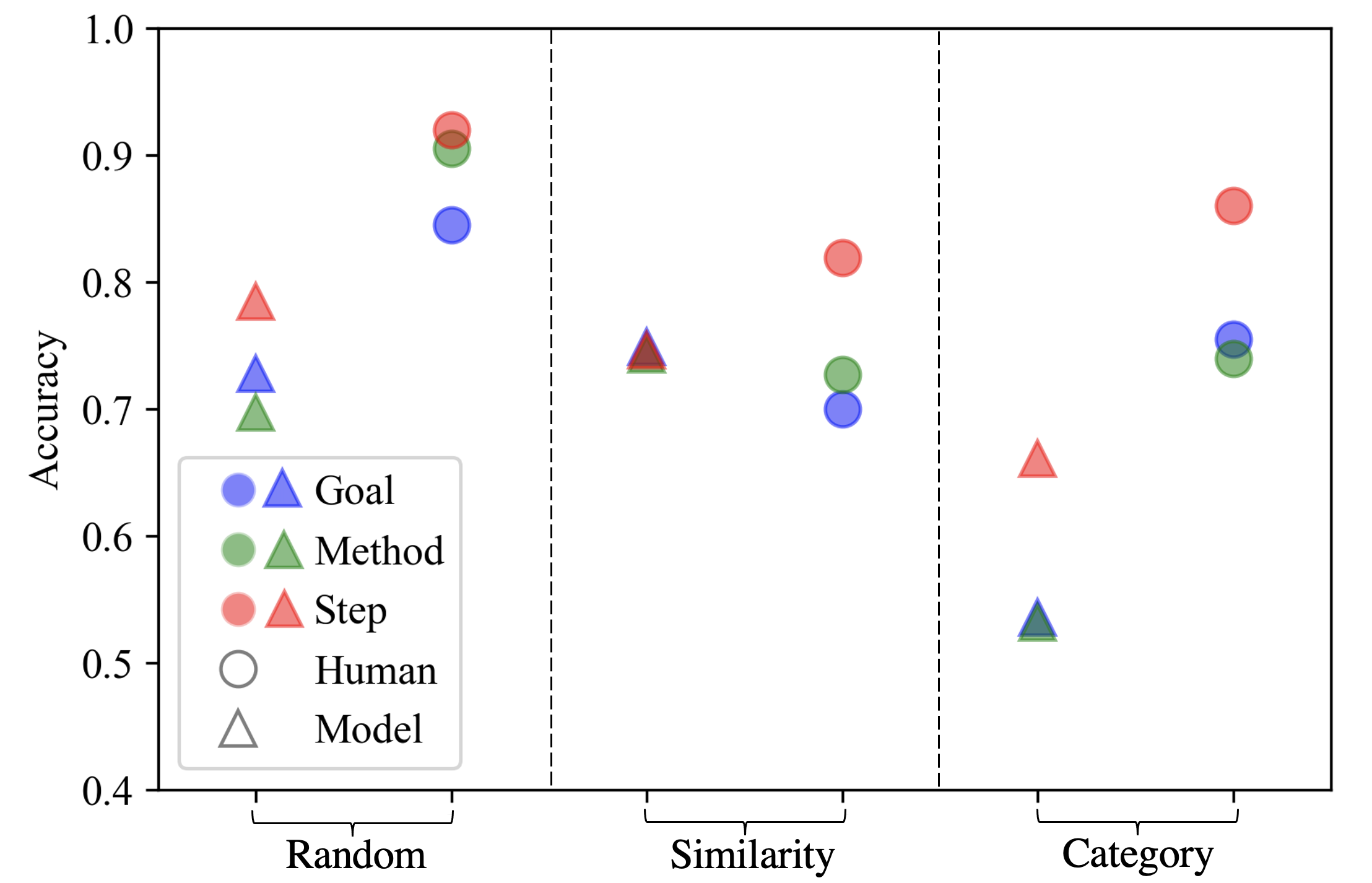}
    \caption{Accuracy of human (circles) and model (triangles) on the modified wikiHow VGSI test set with different textual input (e.g., in Fig~\ref{fig:example}, the \textit{goal} prompt will be replaced by \textit{method} - ``Baking the Fish.'' or \textit{step} - ``Preheat the oven.'').}
    \label{fig:different prompts}
\end{figure}

\section{Experimental Setup}
\subsection{Multiple-Choice Sampling}
We formulate the task as a 4-way multiple choice question, which is easy for evaluating the image-text matching performance and is feasible for human annotation. Specifically, a model is given a textual goal \& four images to predict the most reasonable step towards the goal. We utilize three sampling strategies to obtain negative candidates:

\noindent
\textbf{Random Strategy} We randomly pick three different articles and select one image by chance from each article as the negative sample.

\noindent
\textbf{Similarity Strategy} We greedily select the most similar images based on the feature vectors and use FAISS \cite{johnson2019billion} to retrieve the top-3 most similar images from three different articles.

\noindent
\textbf{Category Strategy} The three negative samples are randomly selected from articles within the same wikiHow category as the prompt goal.

In addition to the multiple-choice format, we also evaluate VGSI in a more realistic goal-image retrieval format (see Appendix~\ref{retrieval}).

\subsection{Human Annotation}
Considering that VGSI is a novel task, we also evaluate how difficult it is for humans. All of our six human annotators are graduate students with good English proficiency. For each annotation test, we selected 100 samples from the testing set. A pair of annotators completed each test and their scores were averaged.

\subsection{Evaluation Metrics}
We report both model and human accuracy for the multiple-choice task. For the retrieval task, we adopt recall at \textit{k} (recall@\textit{k}) and median rank (Med r) to measure the performance (see Appendix~\ref{retrieval}).

\section{Results}
\subsection{In-Domain Results}
Table~\ref{indomain model} shows the performance of the models and humans on the wikiHow dataset. The Triplet Network with BERT embeddings has the best performance. However, there is still a big gap between human and model performance, indicating that VGSI is challenging for even SOTA models. LXMERT performs badly using similarity and category strategies presumably because it heavily depends on grounding objects, and negative samples generated by these two strategies could share similar objects but refer to different goals. Figure~\ref{fig:different prompts} demonstrates that both humans and models perform better with lower-level texts as prompt, which reflects that our VGSI task is more challenging.

\subsection{Transfer Learning}
To robustly show the potential of wikiHow as a multimodal transfer learning resource, we compare it with two existing caption-based datasets, Flickr30K \cite{Plummer_2015_ICCV} and MSCOCO \cite{vinyals2016show}, which are used as pre-training alternatives. We use the official train/val split for each dataset and pre-train two models separately on Flickr and MSCOCO using the same multiple-choice sampling strategies as VGSI.

\begin{table}[!t]
\centering
\resizebox{7.5cm}{!}{%
\begin{tabular}{cc|ccc}
\thickhline
                           &     & \multicolumn{3}{c}{\textbf{Sampling Strategy}}              \\ \hline
\textbf{PT-Data}                    & \textbf{FT?} & \textbf{Random}         & \textbf{Similarity}      & \textbf{Category}        \\ \hline
-                          & \checkmark   &  .6649      & .5085       &   .5216      \\ \hline
\multirow{2}{*}{Flickr30K} & \xmark       & .4903          & .5103          & .3919          \\
                           & \checkmark   & .7006      & .5823         &  .5495         \\ \hline
\multirow{2}{*}{MSCOCO}    & \xmark            & .5349          & .5401          & .4071          \\
                           & \checkmark          & .7481         &  .6180        &  .5536       \\ \hline
\multirow{2}{*}{Howto100m}    & \xmark            & .5694          & .5811          & .3989          \\
                           & \checkmark          & .6948        &  .6104        &  .5436      \\ \hline
\multirow{2}{*}{{\textcolor{outerspace}{\textbf{wiki}}\textcolor{OLIVINE}{\textbf{How}}}}   & \xmark          & \textcolor{RED}{\textbf{.6245}}         & \textcolor{RED}{\textbf{.6309}}           & \textcolor{RED}{\textbf{.4586}}   \\
                           & \checkmark          & \textcolor{BLUE}{\textbf{.7639}}          &  \textcolor{BLUE}{\textbf{.6854}}         &   \textcolor{BLUE}{\textbf{.5659}}       \\ \hline
Human                      &   -  &    .9695       &      .8500           &    .8682       \\ \thickhline
\end{tabular}
}
\caption{Transfer performance (4-way multiple choice accuracy) on COIN. PT stands for pre-training, FT for fine-tuning. FT results are obtained by fine-tuning the model on 5 examples of the COIN training set (i.e., 5-shot). \textcolor{RED}{\textbf{Red}} numbers indicate the best zero-shot performance. \textcolor{BLUE}{\textbf{Blue}} numbers are the best fine-tuned results.} 
\label{COIN_multi}
\end{table}

\begin{table}[!t]
\centering
\resizebox{7.5cm}{!}{%
\begin{tabular}{cc|ccc}
\thickhline
                           &     & \multicolumn{3}{c}{\textbf{Sampling Strategy}}              \\ \hline
\textbf{PT-Data}                    & \textbf{FT?} & \textbf{Random}         & \textbf{Similarity}      & \textbf{Category}        \\ \hline
-                          & \checkmark   & .6005          & .6096          & .4434          \\ \hline
\multirow{2}{*}{Flickr30K} & \xmark       & .4837          & .5398          & .3856          \\
                           & \checkmark            & .6207          & .6408          & .4740          \\ \hline
\multirow{2}{*}{MSCOCO}    & \xmark            & .5099          & .5715          & .3958          \\
                           & \checkmark          & .6340          & .6640          & .4794          \\ \hline
\multirow{2}{*}{COIN}      & \xmark           & .5067          & .5161          & .3978          \\
                           & \checkmark          & .6170          & .6343          & .4638          \\ \hline
\multirow{2}{*}{{\textcolor{outerspace}{\textbf{wiki}}\textcolor{OLIVINE}{\textbf{How}}}}   & \xmark          & \textcolor{RED}{\textbf{.6556}}           & \textcolor{RED}{\textbf{.6754}}           & \textcolor{RED}{\textbf{.4750}} \\
                           & \checkmark          & \textcolor{BLUE}{\textbf{.6855}}          & \textcolor{BLUE}{\textbf{.7249}}          & \textcolor{BLUE}{\textbf{.5143}}          \\ \hline
Human                      &   -  & .8300          &    .7858             & .7550          \\ \thickhline
\end{tabular}
}
\caption{Transfer performance (4-way multiple choice accuracy) on Howto100m. FT results are obtained by fine-tuning the model on the full training set.}
\label{howto_multi}
\vspace{-0.15in}
\end{table}

\subsubsection{Target Datasets \& Keyframe Extraction}
Our transfer targets include COIN and Howto100m, both large-scale datasets of instructional videos. Each video depicts the process of accomplishing a high-level goal, mostly everyday tasks. Since these two datasets are video-based while our task is image-based, we apply a key frame extraction heuristic to get critical frames from videos. We then consider the key frames as steps, thus converting the datasets into the VGSI format.

\noindent
\textbf{Howto100m:} We randomly select 1,000 goals and one video for each goal. To extract key frames, we apply k-means clustering in the feature space of the frames of each video and select the closest frame to each cluster center. We further filter these frames by manually removing unrelated frames such as the introduction, transition animations, repetitive frames, etc. We finally obtain 869 goals\footnote{Some goals have no valid frames remaining after the annotation, and are therefore removed altogether.} with 24.7 frames for each goal.\\
\textbf{COIN:} We randomly select 900 videos (5 videos per goal) to construct the test set, and use the remaining 9,709 videos for training. Since COIN has annotations of textual steps and their corresponding video segment, we randomly select one frame within each video segment as a VGSI candidate, resulting in 230.1 frames per goal.

Then we use these frames to construct the multiple-choice examples with the same three sampling strategies. We also compare using wikiHow against using COIN and Howto100m as pre-training data to perform transfer learning to each other since both are instructional video datasets.

\begin{figure}[!t]
    \includegraphics[width=6.9cm]{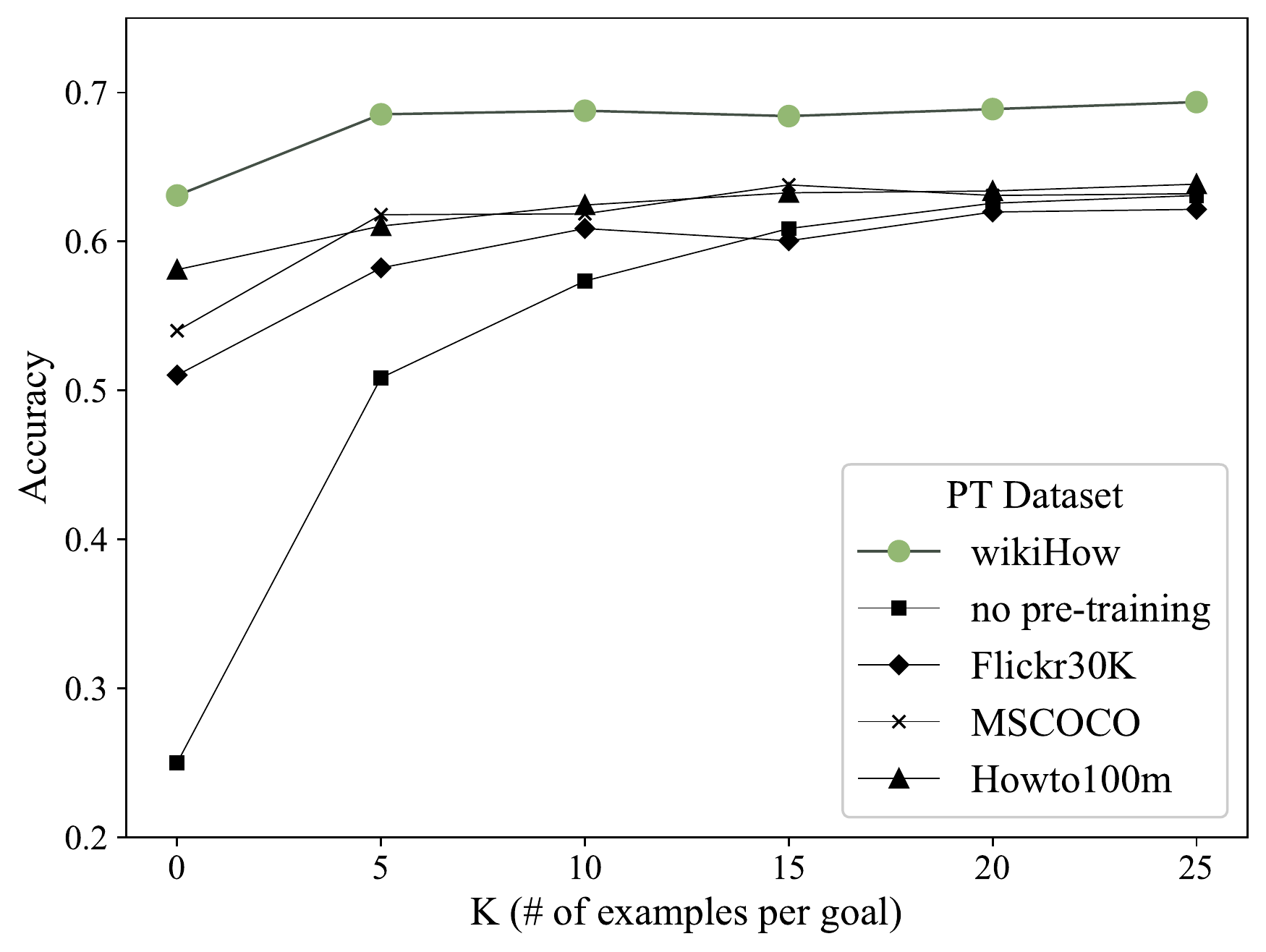}
    \caption{Few-shot performance on COIN (similarity sampling) with different pre-training datasets vs. the number of examples per goal.}
    \label{fig:coin sim}
    \vspace{-0.1in}
\end{figure}

\subsubsection{Transfer Learning Performance}
We use two different transfer learning setups for COIN\footnote{The small number of goals in COIN leads to an extreme imbalance between video frames and texts, which makes it hard for training. Thus there is no train/test split on goals.} and Howto100m. For COIN, we formulate the test as a \textit{K}-shot learning task where \textit{K} is the number of VGSI training examples for each goal. The 180 goals for testing are seen during training to simulate the scenario where we have some instances of a task. For Howto100m, we split the 869 goals into 8:2 for training and testing, where the test goals are unseen during training. 

Tables~\ref{COIN_multi} and~\ref{howto_multi} both indicate that pre-training on wikiHow can improve VGSI performance on out-of-domain datasets. Especially for the Howto100m results, the model pre-trained on wikiHow without fine-tuning outperforms even those pre-trained on other caption-based datasets that were fine-tuned on wikiHow. This is strong evidence that wikiHow can serve as a useful knowledge resource since the learned multimodal representation can be directly applied to other datasets. 

\begin{figure}[!t]
    \includegraphics[width=6.9cm]{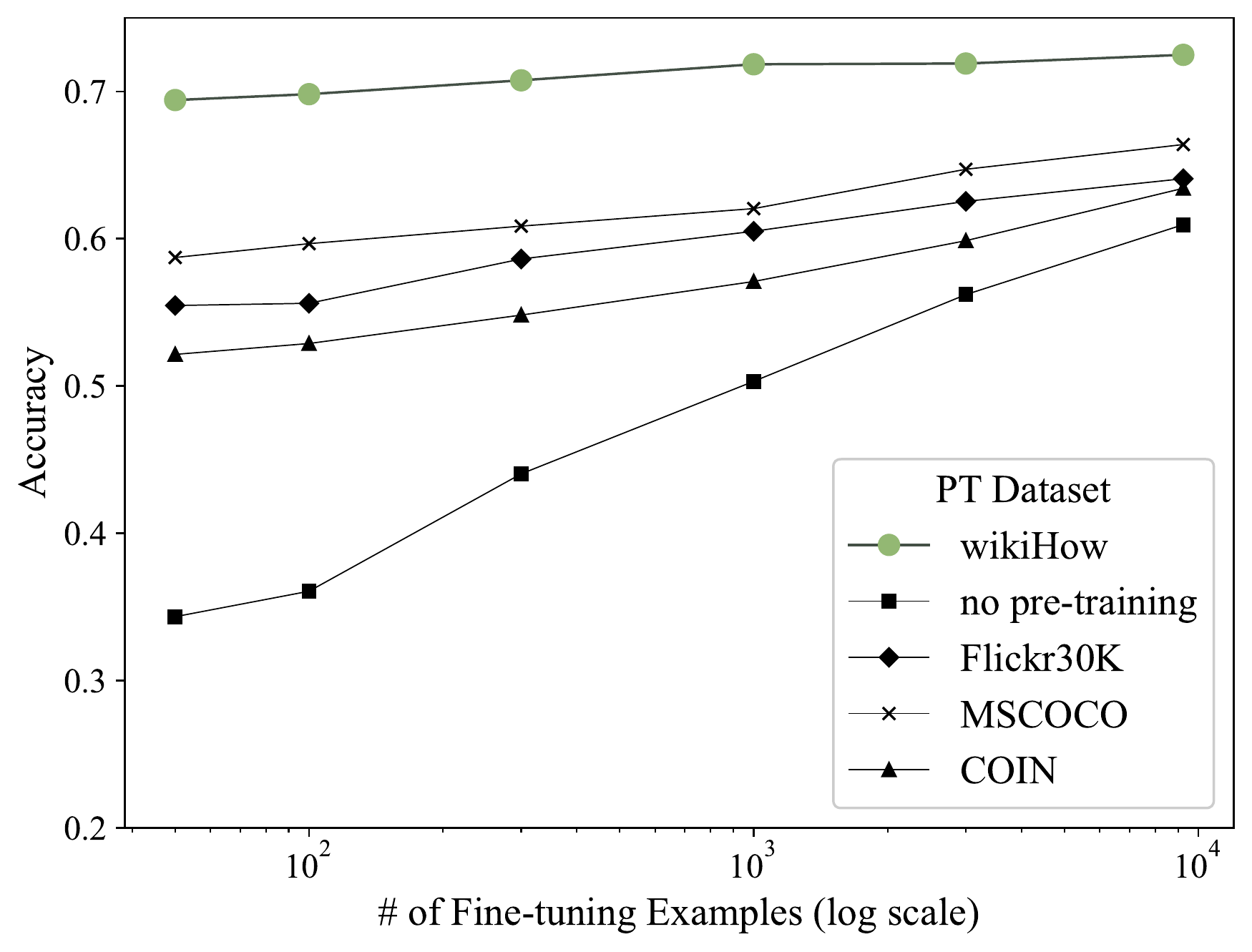}
    \caption{Transfer performance on Howto100m (similarity sampling) with different pre-training datasets vs. the number of training examples.}
    \label{fig:howto random}
    \vspace{-0.1in}
\end{figure}

To further validate whether the advantages of pre-training on wikiHow persist with the increasing number of fine-tuning examples, we report the performance with $K \in \{0, 5, 10, 15, 20, 25\}$ for COIN and training examples ranging from 50 to 9,249 (full) for Howto100m. Shown in Figure~\ref{fig:coin sim} \& ~\ref{fig:howto random}, the model pre-trained on wikiHow consistently outperforms those pre-trained on the other datasets by significant margins with the increase of fine-tuning examples. The curve of wikiHow does not converge with the other curves even with the maximum number of training examples, which reflects that wikiHow could be a reliable pre-training data source for both low- and rich-resource scenarios. 

\section{Conclusion}
In this paper, we propose the novel Visual Goal-Step Inference task (VGSI), a multimodal challenge for reasoning over procedural events. We construct a dataset from wikiHow and show that SOTA multimodal models struggle on it. Based on the transfer learning results on Howto100m and COIN, we validate that the knowledge harvested from our dataset could transfer to other domains. The multimodal representation learned from VGSI has strong potential to be useful for NLP applications such as multimodal dialog systems.

\vspace{-0.1in}
\section*{Acknowledgments}
\vspace{-0.05in}

This research is based upon work supported in part by the DARPA KAIROS Program (contract FA8750-19-2-1004), the DARPA LwLL Program (contract FA8750-19-2-0201), and the IARPA BETTER Program (contract 2019-19051600004). Approved for Public Release, Distribution Unlimited. The views and conclusions contained herein are those of the authors and should not be interpreted as necessarily representing the official policies, either expressed or implied, of DARPA, IARPA, or the U.S. Government.

We thank Chenyu Liu for annotations. We also thank Simmi Mourya, Keren Fuentes, Carl Vondrick, Zsolt Kira, Mohit Bansal, Lara Martin, and anonymous reviewers for their valuable feedback. 

\bibliography{anthology,custom}
\bibliographystyle{acl_natbib}

\clearpage

\newpage

\appendix
\section{Model Implementation Details}
\subsection{Architecture and Loss Function}
\subsubsection{DeViSE}
The Deep-Visual Semantic Embedding (DeViSE) model takes in the pre-trained embedding vectors from two modalities and maps the source vector onto the span of the target vector representation. First, the DeViSE model is only trained on the related (positive) pairs $(G, I)$, and we map the image to the goal ($I \rightarrow G$). Then, the model projects the image embedding onto the same dimension as the goal and we apply L2 normalization to obtain the unit vectors:
\begin{equation} \label{eq2}
\begin{split}
    \hat{X}_I &= L_2N(X_I W_{I \rightarrow G}) \\
    \hat{X}_G &= L_2N(X_G)
\end{split}
\end{equation}
where, $L_2N$ stands for L2 normalization and $W_{I \rightarrow G} \in  \mathbb{R}^{d_I \times d_G}$ is the weight. 

Then the DeViSE model uses a similarity function (here we choose cosine distance) to compute the distance between $\hat{X_I}$ and $\hat{X_G}$ as the loss:
\begin{equation}
\begin{split}
    \mathcal{L}_{DeViSE} &= cos(\hat{X}_I, \hat{X}_G) \\
    match(G, I)_{DeViSE} &= 1 - cos(\hat{X}_I, \hat{X}_G)
\end{split}
\end{equation}
In which $cos$ means the cosine distance. For DeViSE, the matching score is the cosine similarity between the two unit vectors.

\subsubsection{Similarity Network}
A Similarity Network is one type of two-branch networks for matching an image and text. It is a supervised model which takes in $(G_i, I_i, y_i)$, and $y_i \in \{ 0, 1\}$ is the binary label that indicates whether $G_i$ and $I_i$ are related or not.

Each branch of the network maps one modality to the cross-modality span and executes pointwise multiplication to construct a joint vector:
\begin{equation}
    \begin{split}
        \hat{X}_I &= L_2N(X_I W_{I \rightarrow J}) \\
        \hat{X}_G &= L_2N(X_G W_{G \rightarrow J}) \\
        X_J &= \hat{X}_I \odot \hat{X}_G
    \end{split}
\end{equation}
in which, $W_{I \rightarrow J}\in \mathbb{R}^{d_I \times d_J}$ and $ W_{G \rightarrow J} \in \mathbb{R}^{d_G \times d_J}$  are the weights and $\odot$ represents an element-wise product.

The similarity network can be viewed as a binary classifier, and therefore we could use binary cross-entropy (BCE) as the loss function: 

\begin{equation}
\begin{split}
    \mathcal{L}_{sim} =&  - \Sigma_i^N y_i \cdot \log p (y_i) \\
    &+ (1 - y_i) \cdot \log (1 - p(y_i))
\end{split}
\end{equation}

The last layer of the similarity network is a fully-connected layer with a softmax activation function, and the output is an array of size two, in which the elements denote the weight for each class. We compute the matching score by multiplying $+1$ (matched) and $-1$ (unmatched) on these two elements:

\begin{equation}
\begin{split}
    \alpha &= \mathrm{softmax}(\mathrm{fc}(X_J)) \\
    match (G, &I)_{sim} = 1 \cdot \alpha[0] + (-1) \cdot \alpha[1]
\end{split}
\end{equation}

where $\mathrm{fc}$ stands for fully-connected layer.

\subsubsection{Triplet Network}
A Triplet Network requires the input to be in the format of a triplet $(G, I_{pos}, I_{neg})$. There will be three branches in the network which map the three embeddings to the same joint span:
\begin{equation}
    \begin{split}
         \hat{X}_G &= L_2N(X_G W_{G \rightarrow J}) \\
         \hat{X}_{I_{pos}} &= L_2N(X_{I_{pos}} W_{I \rightarrow J}) \\
         \hat{X}_{I_{neg}} &= L_2N(X_{I_{neg}} W_{I \rightarrow J}) \\
    \end{split}
\end{equation}
in which, $W_{G \rightarrow J} \in \mathbb{R}^{d_G \times d_J}$ and $W_{I \rightarrow J} \in \mathbb{R}^{d_I \times d_J}$ are weights, and the branches of positive and negative images share the same weight.

The network learns the cross-modality by minimizing the distance between positive pairs and maximizing the distance between negative pairs. We choose cosine distance as the distance function which will also be used to compute the matching score:
\begin{equation}
    \begin{split}
        \mathcal{L}_{trip} = &\mathrm{max} (0, cos(\hat{X}_G, \hat{X}_{I_{pos}}) \\
        &- cos(\hat{X}_G, \hat{X}_{I_{neg}}) + \mathrm{m}) \\
        mat&ch(G, I)_{trip} = cos(\hat{X}_G, \hat{X}_I)
    \end{split}
\end{equation}
Where m is the margin, which is set to $0.2$ in the experiment.

\subsubsection{LXMERT}
LXMERT \cite{tan2019lxmert} is a multimodal encoder that aims to ground text to images. It takes as an input image $I$ and a related sentence $G = \{w_1, w_2, \ldots, w_n\}$. The image objects are embedded using a feature extractor \cite{Anderson2017up-down} pre-trained on ImageNet \cite{deng2009imagenet}. Given $I$ the detector finds $m$ objects $\{o_1, o_2, \ldots, o_m\}$ where: $o_i = \{p_i, f_i\}$, s.t. $p_i$ is its bounding box and $f_i$ is its 2048-dimensional region of interest (RoI). LXMERT learns a position-aware embedding as follows:  
\begin{align}
    f'_i &= L_2N(W_Ff_i + b_F) \\
    p'_i &= L_2N(W_Pp_i + b_P) \\
    v_i &= ( f'_i+ p'_i)/ 2
\end{align}

The text tokens are extracted using a tokenizer \cite{wu2016google} and converted to index-aware embeddings s.t. $w_i$ and $i$ are projected onto embedding spaces $w'_i$, $u'_i$, to get a common embedding. 
\begin{align}
    h_i = L_2N(w'_i+u'_i)
\end{align} 

Those inputs are then passed through a language encoder $E_G$, an object relationship encoder $E_I$, and a cross-modality transformer encoder $E_J$. Let $X_I = \{v_1, v_2, \ldots, v_n\}$ and $X_G = \{h_1, h_2, \ldots, h_n\}$.    
\begin{equation}
\begin{split}
    \hat{X_G} &= E_G(X_G) \\
    \hat{X_I} &= E_I(X_I) \\
    X_{G_J}, X_{G_I} &= E_J(\hat{X_G}, \hat{X_I})
\end{split}
\end{equation}
Then the cross-modality output $X_J$ is extracted from the output embedding $X_{G_J}$ that corresponds to the special token [CLS] appended to each input text. 

Similarly to A.1.2, we use BCE loss.

\begin{equation}
\begin{split}
    \mathcal{L}_{lxmert} =&  - \Sigma_i^N y_i \cdot \log p (y_i) \\ 
    &+ (1 - y_i) \cdot \log (1 - p(y_i))
\end{split}
\end{equation}

and compute the matching score:
\begin{equation}
\begin{split}
    \alpha = &\mathrm{softmax}(\mathrm{fc}_2\mathrm{fc}_1(X_J)) \\
    match (G, I)_{lxmert} &= 1 \cdot \alpha[0] + (-1) \cdot \alpha[1]
\end{split}
\end{equation}
\subsection{Features}
\subsubsection{Vision}
We select InceptionV3 \cite{DBLP:journals/corr/SzegedyVISW15} as the feature extractor for the image. We have tried VGG19 and Resnet50, but InceptionV3 turns out to have the best performance. We use the second last hidden layer of InceptionV3 to obtain a vector of (2048, ). 

\subsubsection{Language}
We use a pre-trained BERT sentence transformer \cite{reimers-2019-sentence-bert} with \texttt{bert-base-uncased} as our base model. Then, we use max-pooling to get the feature vector with a dimension of (768, ).

\subsection{Hyper Parameters}
See Table~\ref{hyper}.
\begin{table}[!t]
\centering
\resizebox{7.5cm}{!}{%
\begin{tabular}{c|cccc}
\thickhline
      \textbf{Model}        & \textbf{Optimizer} & \begin{tabular}[c]{@{}c@{}}\textbf{Learning} \\ \textbf{Rate}\end{tabular} & \begin{tabular}[c]{@{}c@{}}\textbf{Batch} \\ \textbf{Size}\end{tabular} & \begin{tabular}[c]{@{}c@{}} \textbf{n. of} \\ \textbf{Parameters} \end{tabular} \\ \hline
DeViSE         & RMSProp   & 5e-6                                                     & 1024                                                  & 2,897,664                                                  \\
Similarity Net & RMSProp   & 5e-6                                                     & 1024                                                  & 4,424,170                                                  \\
Triplet Net    & Adam      & 1e-5                                                     & 1024                                                  & 4,984,832                                                  \\
LXMERT         & Adam      & 5e-7                                                         &  32                                    &                              209,124,098                             \\ \thickhline
\end{tabular}
}
\caption{Hyper Parameters of All Models.}
\label{hyper}
\end{table}

\subsection{Training Details}
The training of DeViSE, Similarity Network and Triplet Network were on a single NVIDIA RTX 2080 for 200 epochs with early stopping. The training took less than 10 hours.

We used a pre-trained LXMERT model with 9 language layers, 5 cross-encoder layers, 5 vision encoder layers, and a 2 layer linear classification head, with \verb GELU() \cite{hendrycks2016gaussian}  and \verb ReLU()  activation, with a Sigmoid final layer and with  normalization in the first layer. 

We fine-tune the model for 10 epochs while allowing the gradient to flow through the LXMERT pre-trained layers. We use a binary cross-entropy loss from the PyTorch library and an Adam \cite{kingma2014adam} optimizer. Note that we deal with imbalanced datasets by repeating the positive samples and shuffling the data.

\section{Goal-Image Retrieval Task} \label{retrieval}
\subsection{Sampling}
Goal-Image Retrieval is a more practical format that gives a high-level goal and a pool of images and aims to rank these images based on their similarity with the goal query. 

In this experiment, we randomly select 1,000 high-level goals from the testing set of multiple-choice tasks and choose 5 images for each goal, thus building a pool of 5,000 images.

\subsection{Evaluation Metrics}
We perform recall at \textit{k} (recall@\textit{k}, higher the better) and median rank (Med r, lower the better) to measure the retrial performance. For the 5k image pool, $\textit{k} \in \{ 10, 25, 50, 100\}$, while for the 1k image pool, $\textit{k} \in \{1, 5, 10, 25\}$.

\subsection{In-Domain Performance}
As shown in Table~\ref{model_retrieve}, the triplet network with BERT as the text embedding has the best performance.
\begin{table}[!t]
\centering
\resizebox{7.5cm}{!}{
\begin{tabular}{c|ccccc}
\thickhline
\multirow{2}{*}{\textbf{Model}} & \multicolumn{5}{c}{\textbf{5K Testing Images}} \\ \cline{2-6} 
                       & \textbf{R@10}  & \textbf{R@25}  & \textbf{R@50}  & \textbf{R@100} & \textbf{Med r} \\ \hline
Random                 & 0.1  & 0.4  & 1.0  & 2.1  & 2519  \\
DeViSE                 & 5.2  & 9.8  & 15.2 & 23.8 & 429   \\
Similarity Net         & 5.8  & 11.5 & 17.6 & 27.0 & 347   \\
Triplet Net (GloVe)    & 5.9  & 12.2 & 19.9 & 31.2 & 264   \\
Triplet Net (BERT)     & \textbf{6.9}  & \textbf{13.8} & \textbf{21.9} & \textbf{32.7} & \textbf{249}   \\ \thickhline
\end{tabular}
}
\caption{In-Domain Retrieval results with Different Models.}
\label{model_retrieve}
\end{table}

\subsection{Query on Different Prompts}
As can be seen from Table~\ref{prompt_retrieve}, the model has higher performance when using the detailed step description as a prompt. Through qualitative analysis (see Figure~\ref{fig:qual_examples_2}) on some samples, we discovered that some method descriptions are very general, and short abstract keywords are even more refined than the goal description. To quantify this finding, we calculate the average length of tokens (remove stop words) and the vocabulary size of the three types of prompts. Apparently, the step description is more fruitful than the method and goal with higher token length and vocab size. The method described has a lower average length of tokens, which is in line with our observation.

\begin{table}[!t]
\resizebox{7.5cm}{!}{%
\begin{tabular}{c|cc|ccccc}
\thickhline
\multirow{2}{*}{\textbf{Prompt}} & \multirow{2}{*}{\begin{tabular}[c]{@{}c@{}} \textbf{Token}\\ \textbf{Length}\end{tabular}} & \multirow{2}{*}{\begin{tabular}[c]{@{}c@{}}\textbf{Vocab}\\ \textbf{Size}\end{tabular}} & \multicolumn{5}{c}{\textbf{1K Testing Images}}                                     \\ \cline{4-8} 
                        &                                                                                &                                                                       & \textbf{R@1}         & \textbf{R@5}          & \textbf{R@10}         & \textbf{R@25}         & \textbf{Med r}      \\ \hline
Goal                    & 3.34                                                                           & 19,299                                                                & 4.6          & 14.6          & 22.8          & 36.3          & 49          \\
Method                  & 3.11                                                                           & 24,180                                                                & 2.5          & 11.4          & 18.9          & 33.3          & 57          \\
Step                    & \textbf{4.67}                                                                           & \textbf{49,999}                                                              & \textbf{6.1} & \textbf{20.1} & \textbf{31.4} & \textbf{48.7} & \textbf{26} \\ \thickhline
\end{tabular}
}
\caption{Query on Different Prompts}
\label{prompt_retrieve}
\end{table}

\subsection{Transfer Performance on Retrieval}
We also evaluate the transfer performance on a retrieval task. For COIN, we choose 5-6 images for each video from the 180 goals and construct a pool of 1,000 images. For Howto100m, we randomly select 5-6 images of each of the videos in the testing set and also form a pool of 1K images. 

Table~\ref{coin_retri} and~\ref{howto_retri} indicates the model pre-trained on wikiHow outperforms the other dataset in the retrieval task and the aggregation model could further improve the performance. 

\section{Step-Aggregation Model} \label{step_agg}

\begin{table}[!t]
\resizebox{7.5cm}{!}{%
\begin{tabular}{c|ccccc}
\thickhline
               & \multicolumn{5}{c}{\textbf{1k Test Images}} \\ \hline
\textbf{PT-Data}        & \textbf{R@1}  & \textbf{R@5}  & \textbf{R@10}  & \textbf{R@25}  & \textbf{Med r} \\ \hline
-              & 0.0 & 0.5 & 1.2  & 2.5  & 517   \\
Flickr         & 1.2 & 4.0 & 7.1  & 14.2 & 240   \\
MSCOCO         & 0.9 & 5.5 & 9.3  & 19.0 & 170   \\
{\textcolor{outerspace}{\textbf{wiki}}\textcolor{OLIVINE}{\textbf{How}}}     & \textbf{1.4} & \textbf{7.6} & \textbf{12.6} & \textbf{23.8} & \textbf{102}   \\ \thickhline
\end{tabular}
}
\caption{Zero-shot Retrieval on COIN}
\label{coin_retri}
\end{table}

\begin{table}[!t]
\resizebox{7.5cm}{!}{%
\begin{tabular}{cc|ccccc}
\thickhline
                           &                           & \multicolumn{5}{c}{\textbf{1K Test Images}} \\ \hline
\textbf{PT-Data}                    & \textbf{FT?}                       & \textbf{R@1}  & \textbf{R@5}  & \textbf{R@10}  & \textbf{R@25}  & \textbf{Med r} \\ \hline
-                          & \checkmark & 1.1  & 4.4  & 9.5   & 17.4  & 129   \\ \hline
\multirow{2}{*}{Flickr30K} & \xmark     & 0.9  & 3.9  & 6.9   & 11.7  & 213   \\
                           & \checkmark & 1.2  & 5.4  & 10.5  & 20.9  & 122   \\ \hline
\multirow{2}{*}{MSCOCO}    & \xmark     & 0.5  & 4.1  & 7.3   & 13.8  & 202   \\
                           & \checkmark & 1.7  & 6.8  & 11.9  & 22.4  & 98    \\ \hline
\multirow{2}{*}{COIN}      & \xmark     & 1.1  & 4.2  & 7.7   & 15.3  & 193   \\
                           & \checkmark & 1.6  & 6.1  & 11.7  & 21.6  & 118   \\ \hline
\multirow{2}{*}{{\textcolor{outerspace}{\textbf{wiki}}\textcolor{OLIVINE}{\textbf{How}}}}   & \xmark     & \textcolor{BLUE}{\textbf{1.6}}  & \textcolor{BLUE}{\textbf{7.3}}  & \textcolor{BLUE}{\textbf{13.5}}  & \textcolor{BLUE}{\textbf{25.1}}  & \textcolor{BLUE}{\textbf{88}}    \\
                           & \checkmark & \textcolor{RED}{\textbf{2.0}}  & \textcolor{RED}{\textbf{7.9}}  & \textcolor{RED}{\textbf{14.7}}  & \textcolor{RED}{\textbf{26.7}}  & \textcolor{RED}{\textbf{84}}    \\ \thickhline
\end{tabular}}

\caption{Retrieval Performance on Howto100m}
\label{howto_retri}
\end{table}

We have seen that SOTA models do not perform well in VGSI because of the implicit vision-language relation. So we develop a step aggregation model that takes advantage of the existing goal-step knowledge from wikiHow. The main idea is as follows: given an unseen textual goal, we use \textit{k}-nearest neighbors to find the most related article title from wikiHow, then extract the $n$ steps from this article as $S = \{s_1, s_2, ..., s_n\}$. Instead of directly using the given goal to match the images (goal score - $\mathrm{Score}_{g}$), we could use the sequence of steps to improve the matching (step score - $\mathrm{Score}_{s}$). Then use linear interpolation to summarize these two scores as our final matching score.

\begin{equation}
\begin{split}
    &\mathrm{Score}_{g} = match(G, I) \\
    &\mathrm{Score}_{s} = \mathrm{max}_{i = 1:n}(match(s_i, I)) \\
    &\mathrm{Score}_{final} = \lambda \cdot \mathrm{Score}_{g} + (1 - \lambda) \cdot \mathrm{Score}_{s}
\end{split}
\end{equation}
where, $\lambda$ adjusts the step and goal scores weights, we choose $\lambda = 0.5$.

\begin{figure*}[!t]
\centering
    \includegraphics[width=15cm]{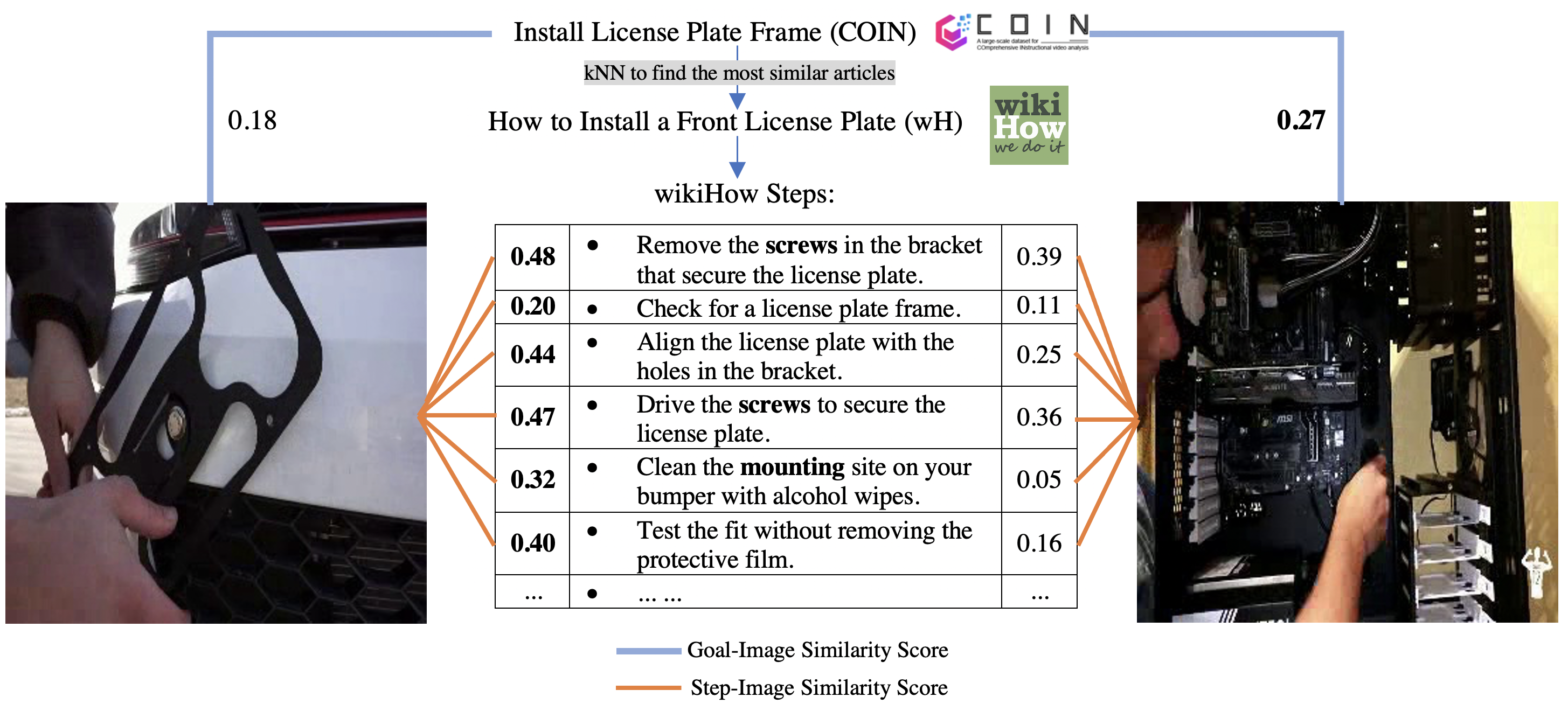}
    \caption{The architecture of the Step-Aggregation Model.}
    \label{fig:aggregation model}
\end{figure*}

\begin{table*}[!t]

\centering
\begin{tabular}{c|c|ccc}
\thickhline
\multirow{2}{*}{\textbf{Dataset}}   & \multirow{2}{*}{\textbf{Model}}           & \multicolumn{3}{c}{\textbf{Sampling Strategy}}        \\ \cline{3-5} 
                           &                                  & \textbf{Random}        & \textbf{Similarity}    & \textbf{Category}      \\ \hline
\multirow{2}{*}{COIN}      & wikiHow                          & .7639         & .6854         & .5659         \\
                           & wikiHow$^{agg}$ & .\textbf{7657}(+0.2\%) & \textbf{.6942}(+1.3\%) & \textbf{.5764}(\textbf{+1.9\%}) \\ \hline \hline
\multirow{2}{*}{Howto100m} & wikiHow                          & .6855         & .7249         & .5143         \\
                           & wikiHow$^{agg}$ & \textbf{.6947}(+1.3\%) & \textbf{.7392}(\textbf{+2.0\%}) & \textbf{.5245}(\textbf{+2.0\%}) \\ \thickhline
\end{tabular}
\caption{Apply Step-Aggregation model on multiple-choice VGSI ($agg$ stands for aggregation model).}
\label{agg multi}
\end{table*}

\begin{table*}[!t]
\begin{tabular}{c|c|ccccc}
\thickhline
\textbf{Dataset}                    & \textbf{Model}   & \textbf{R@1}          & \textbf{R@5}         & \textbf{R@10}         & \multicolumn{1}{l}{\textbf{R@25}} & \multicolumn{1}{l}{\textbf{Med r}} \\ \hline
\multirow{2}{*}{COIN}      & wikiHow & 1.4          & 7.6         & 12.6         & 23.8                     & 102                       \\
                           & wikiHow$^{agg}$ & \textbf{1.9}(+35.7\%) & \textbf{7.8}(+2.6\%) & \textbf{13.6}(+7.9\%) & \textbf{25.9}(+8.8\%)             & \textbf{97}(-4.9\%)                \\ \hline \hline
\multirow{2}{*}{Howto100m} & wikiHow & 2.0          & 7.9         & 14.7         & 26.7                     & 84                        \\
                           & wikiHow$^{agg}$ & \textbf{2.1}(+5.0\%)  & \textbf{8.3}(+5.1\%) & \textbf{15.8}(+7.5\%) & \textbf{27.7}(+3.7\%)             & \textbf{80}(-4.8\%)                \\ \thickhline
\end{tabular}
\caption{Apply Step-Aggregation model on retrieval VGSI.}
\label{agg retri}
\end{table*}

The main idea of the model is to break down the high-level goal into intermediate steps via schema. Then we use the induced sequence of steps as the new query to improve the matching performance. For example in Figure~\ref{fig:aggregation model}, when we want to match the goal “Install License Plate” with two images, the model makes a wrong choice because the negative sample (the right one) also involves the "install" action. However, we could fetch the intermediate steps from \textcolor{outerspace}{wiki}\textcolor{OLIVINE}{How} and use these steps to match the images. The left image (the correct choice) has a higher Step-Image similarity score than the right one. Therefore, the model could improve its performance with the help of this step information. As we can see from the example steps, they contain some useful entities such as “screw”, “bracket”, “bumper”, etc., which are closely related to the visual information in the image but do not show up in the goal sentence.

We apply the aggregation model on both multiple-choice and retrieval VGSI tasks. As shown in Table~\ref{agg multi} and~\ref{agg retri}, with the assistance of the aggregation model, the accuracy of multiple-choice increased by 0.2\% - 2\%, and the median rank of retrieval decreased by 5\%. Since our approach to utilize these steps is very simple, but still achieve a marginal improvement. We hope to see more advanced models to realize the full potential of wikiHow steps.

\begin{figure*}[!t]
\centering
    \includegraphics[width=15cm]{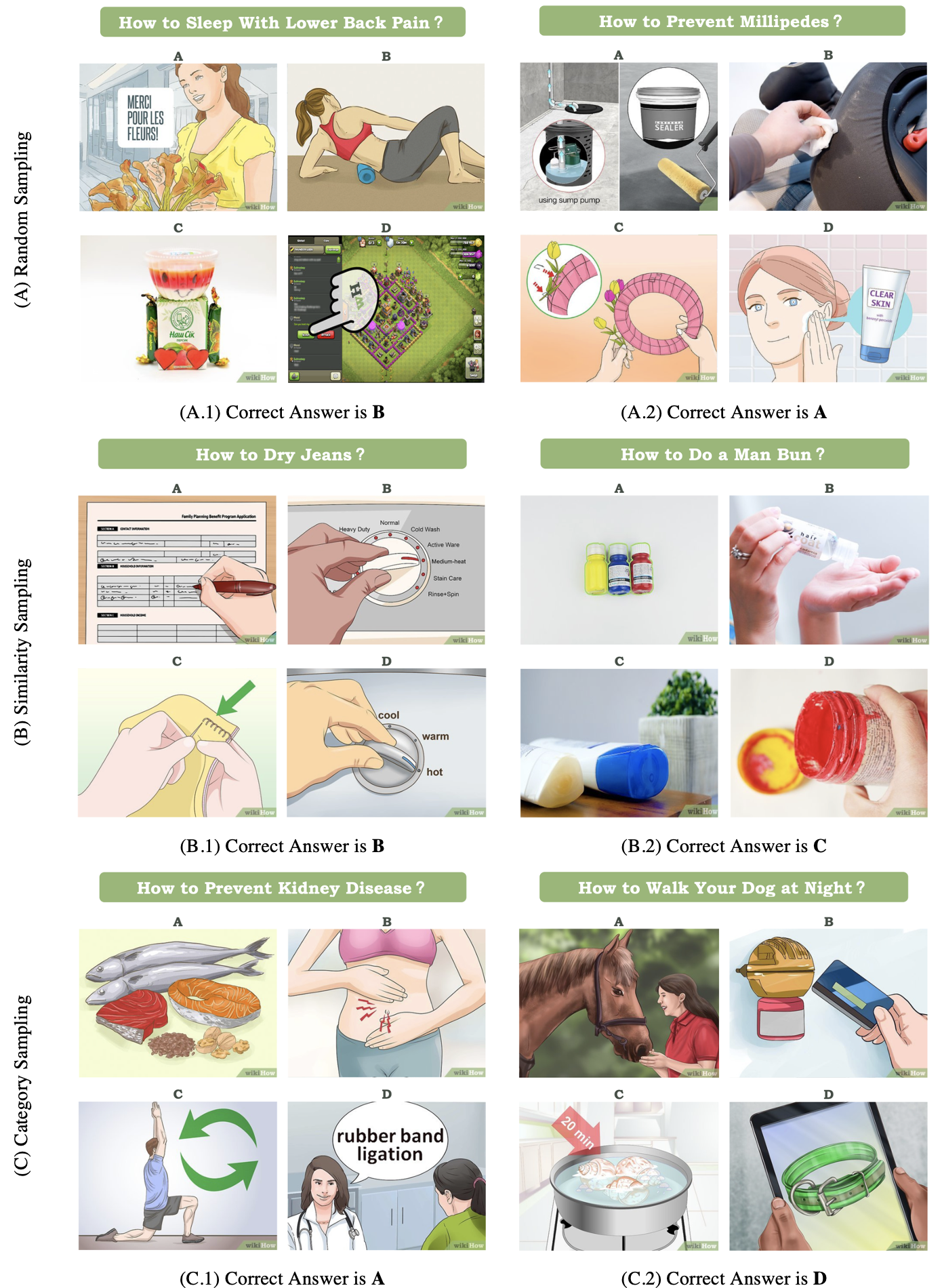}
    \captionsetup{justification=centering}
    \caption[model]{Qualitative Examples Using Different Sampling Strategies.}
    \label{fig:qual_examples_1}
\end{figure*}

\newpage

\section{Qualitative Examples}

\begin{figure*}[!t]
\centering
    \includegraphics[width=15cm]{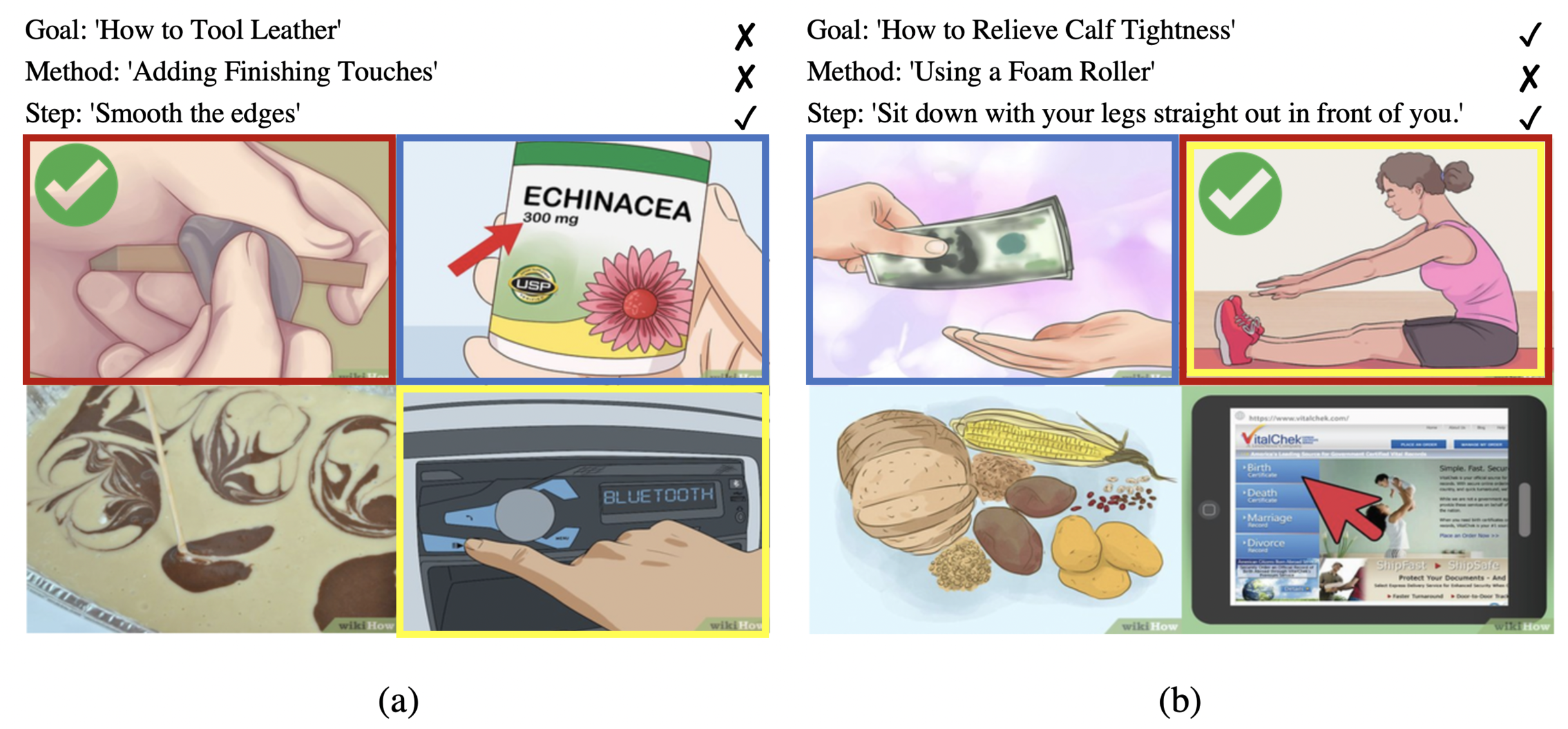}
    \caption[model]{Qualitative Examples Using Different Query Prompts. (Yellow bounding box is the goal's prediction, blue bounding box denotes the method's prediction, red bounding box denotes the step's prediction, green checkmark represents the ground truth.)}
    \label{fig:qual_examples_2}
\end{figure*}

\begin{figure*}[!t]
\centering
    \includegraphics[width=15cm]{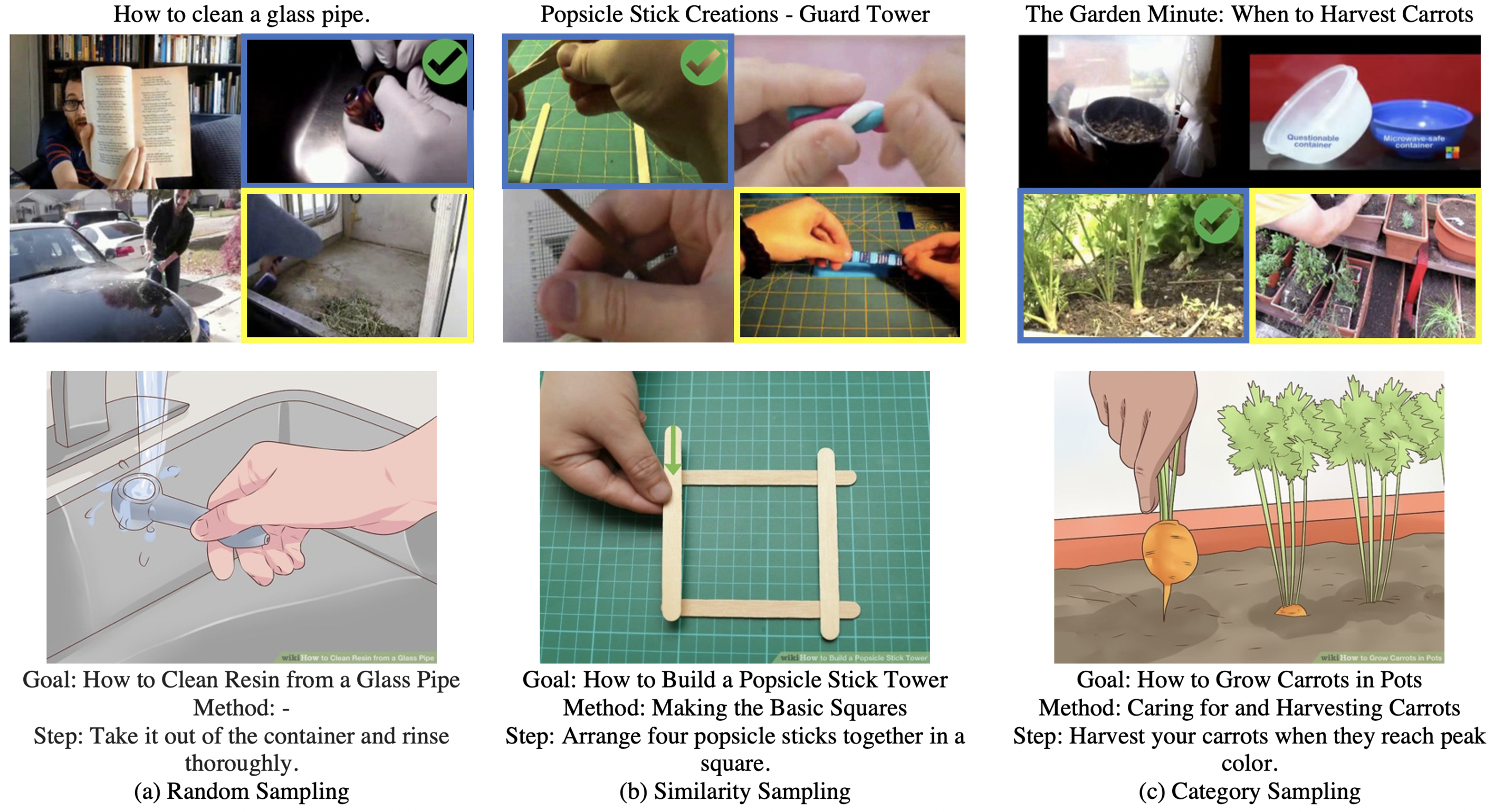}
    \caption[model]{Qualitative Examples of Transfer Learning on Howto100m. 
    (The first row shows the multiple-choice examples of Howto100m video frames, the yellow bounding box is the prediction of the model without pre-training on wikiHow, blue bounding box denotes the prediction of the pre-trained model, and green checkmark represents the ground truth. The second row shows the related images and descriptions we found in wikiHow.)}
    \label{fig:qual_examples_3}
\end{figure*}

See Figure~\ref{fig:qual_examples_1},~\ref{fig:qual_examples_2},~\ref{fig:qual_examples_3}.

\end{document}